\documentclass[sigconf]{acmart}
\usepackage{booktabs} 
\usepackage{algorithm}
\usepackage{algorithmic}
\usepackage{times}
\usepackage{graphics}
\usepackage{graphicx}
\usepackage{amsmath}
\usepackage{amssymb}
\usepackage{xcolor}
\usepackage{multirow}
\usepackage{geometry}
\usepackage{gensymb}
\usepackage[]{hyperref}

\usepackage[normalem]{ulem}

\acmConference[SIGSPATIAL'17]{International Conference on Advances in Geographic Information Systems}{November 2017}{Los Angeles, California, USA}
\acmYear{2017}

\begin{document}
\title{Lane Boundary Geometry Extraction from Satellite Imagery}

\author{Andi Zang}
\affiliation{%
  \institution{Northwestern University}
}
\email{andi.zang@u.northwestern.edu}

\author{Runsheng Xu}
\affiliation{%
  \institution{Northwestern University}
}
\email{runshengxu2017@u.northwestern.edu}

\author{Zichen Li}
\affiliation{%
  \institution{New York University}
}
\email{zichenli@nyu.edu}

\author{David Doria}
\affiliation{%
  \institution{HERE Company}
}
\email{david.doria@here.com}

\begin{abstract}

Automated driving is becoming a reality where \textit{HD Map} plays an important role in path planning and vehicle localization. Lane boundary geometry is one of the key components of HD Map. It is typically created from ground level LiDAR and imagery data, which have their limitations such as prohibitive cost, infrequent update, traffic occlusions, and incomplete coverage. In this paper, we propose a novel method to automatically extract lane information from satellite imagery using pixel-wise segmentation, which addresses the aforementioned limitations. We will also publish a dataset consists of satellite imagery and the corresponding lane boundaries as ground truth to train, test, and evaluate our method.
\end{abstract}

\keywords{HD Maps, lane boundary geometry, satellite image processing}

\maketitle
\section{Introduction and Motivation}

\begin{figure}[H]
	\begin{tabular}{ccc}
		\resizebox{0.15\textwidth}{!}{\rotatebox{0}{
				\includegraphics{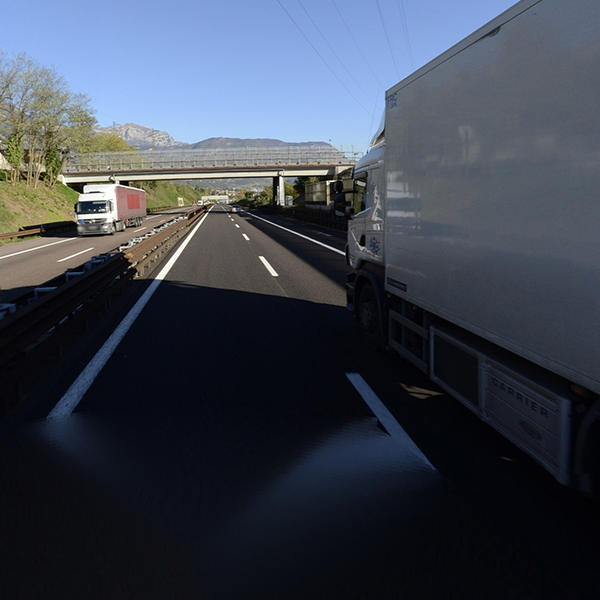}}}
		&
		\resizebox{0.15\textwidth}{!}{\rotatebox{0}{
				\includegraphics{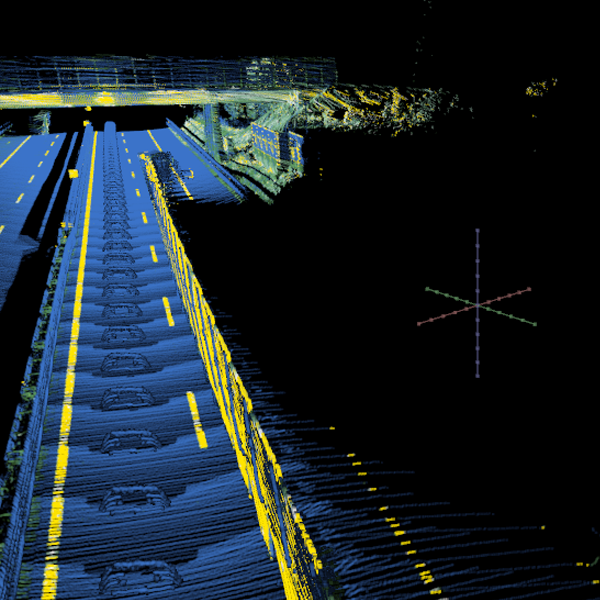}}}
		&
		\resizebox{0.15\textwidth}{!}{\rotatebox{0}{
				\includegraphics{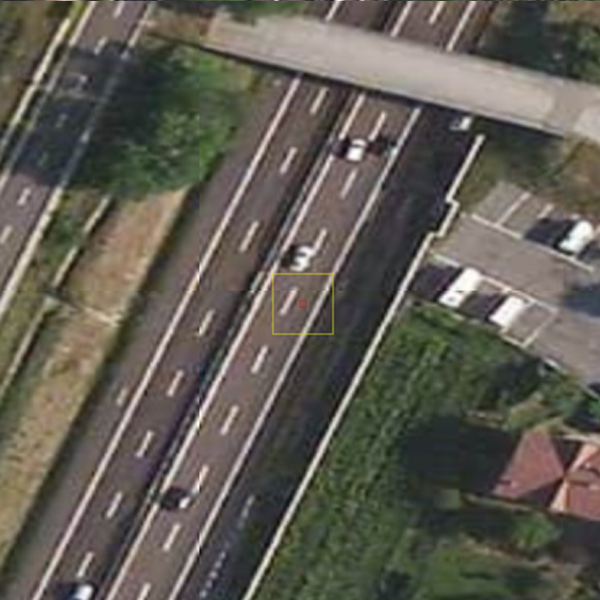}}}
		\\
		(a) & (b) & (c)
		\\
	\end{tabular}
	\caption{Object occlusion in perspective view (a), a wall caused by occlusion in LIDAR point cloud (b), and satellite image (c).}
	\label{fig:occlusion}
\end{figure}

Highly accurate, precise, and detailed lane-level maps, as described in Open Lane Model by the Navigation Data Standard~\cite{nds2016olm}, is critical in enhancing driving safety and empowering automated driving. Lane-level maps enhances vehicle sensor functionality for contextual analysis of the environment, assist vehicle in executing controlled maneuvers beyond sensing range, and provide precise vehicle positioning and orientation on HD Maps. It works in conjunction with signs, barriers, poles, and surface markings to provide the vehicle a more comprehensive and accurate knowledge of the environment. One of the most important attribute in lane-level maps is lane boundary and geometry. Given millions of kilometers of roads in the world, it is cost-prohibitive and time-consuming to manually create and maintain such lane information at a centimeter-level. Ground-level imagery and LiDAR are two of the primary data sources to automatically extract lane information~\cite{hillel2014recent}. ~\cite{broggi1995vision} proposed the method to detect roads from airborne imagery using color and edge information; ~ \cite{yang2013semi,pazhayampallil2013deep,huval2015empirical} proposed to detect road surface and lane markings from LiDAR using the highly accurate and precise 3D measurements in a LiDAR point cloud. Moreover, point cloud aligned with perspective imagery can generate training data~\cite{gu2015fusion} to assist lane-marking detection in perspective imagery. Object occlusion is one of the biggest challenge of road/lane extraction in both LIDAR point cloud and ground-level imagery. As an example, in Figure~\ref{fig:occlusion}, a truck that drives along side the data acquisition vehicle at a similar speed casts a \textbf{"wall"} in the middle of the road. 

Recently, a lot of work has been done to automate lane-level map generation using vehicle-sensor meta data crowd-sourced from large fleet of vehicles~\cite{chen2016detecting} in addition to ground level data such as imagery~\cite{chen2014determining, chen2015detecting1, chen2015detecting2}, LiDAR~\cite{zavodny2016method}, GPS, and Inertial Measurement Unit(IMU) collected by mobile mapping vehicles.

Extracting road/lane information from airborne imagery has its advantage over terrestrial data due to its comprehensive coverage, low-cost, ease to update. The history of road extraction from orthogonal imagery (e.g. satellite and aerial) can be traced back to more than forty years ago; however, limited by image resolution (typically over 2 meters per pixel), traditional approaches rely on edge detection, color segmentation, linear feature detection, and topological linking~\cite{trinder1998automatic,bajcsy1976computer} to extract road networks from overhead imagery. In recent years, more machine-learning based approaches are proposed to detect patch/pixel-wise road region~\cite{mnih2010learning, heitz2008learning, chaudhuri2012semi, mokhtarzade2007road, hu2007road}. These road centric approaches still cannot model the lane-level features even though the common satellite image resolution has improved to 0.5 meter per pixel.

Now, satellite imagery can have a resolution of 0.5 meter per pixel or less, which allows us to utilize the classic approaches with much detailed imagery to model lane-level features~\cite{mattyus2015enhancing, mattyus2016hd, seo2012ortho, pink2010automated}. There are still two challenges after the lane boundary line is successfully detected: the representation of the lane model and the evaluation metric of model accuracy and performance. In paper~\cite{pink2010automated}, the road model is represented as a collection of unstructured lines without attributes; while in paper~\cite{seo2012ortho}, the definition of accuracy is based on the percentage of pixel-wise overlap comparing to their manually drawn line masks in the input images. Hence, their claimed accuracy is less persuasive.



Autonomous vehicles are becoming more of a reality. The increasing demand of HD mapping can be predicted, especially for interstate transportation (i.e. autonomous truck~\cite{autotruck}. The three largest highway networks in the world, U.S., China, and India, are $103$, $446$, and $79$ thousand kilometers~\cite{dot,roadtraffic} long, respectively, which motivates us to concentrate on highway-level road network in this paper. We propose a novel, automated lane boundary extraction technique from satellite imagery. Our approach consists of two stages: pixel-wise line segmentation and hypotheses-grouping classification linking. The pixel-wise line segmentation approach contains patch-based lane-marking classification, and for each positive patch, we segment line pixels to generate line candidates. Hypotheses-linking connects each line candidate by minimizing the proposed cost function to generate structured lane model. A formalized road-model-accuracy-metric is designed to evaluate the results rigorously. We also manually extracted lane boundary ground truth from our dataset. Along with satellite imagery, it can be used for training, testing, and evaluation for comparative studies.

\section{Methodology}
Our lane-boundary-geometry extraction approach contains two stages. In training, similar to~\cite{gu2015fusion}, we use the ground truth lane boundary geometry and the corresponding satellite imagery from Bing Tile Server as input, project lane boundary lines to imagery, crop image into small patches, and train our patch level classifier. In extraction, our approach uses pre-trained classifier, target route/trajectory, and corresponding satellite imagery as input, detects patch-level lane marking candidates~\cite{costea2016aerial}, segment the pixel-wise lane marking candidates, and links~\cite{seo2012ortho} the pixel-wise candidates to generate lane boundary geometry.

\subsection{Patch and Patch Level Classification}\label{tile_and_patch}
\begin{figure}[h]
	\centerline{
	\begin{tabular}{cccc}
		\resizebox{0.12\textwidth}{!}{\rotatebox{0}{
				\includegraphics{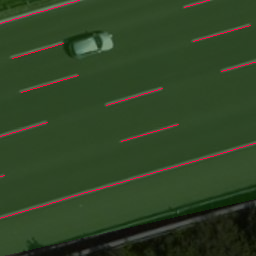}}}
		\resizebox{0.12\textwidth}{!}{\rotatebox{0}{
				\includegraphics{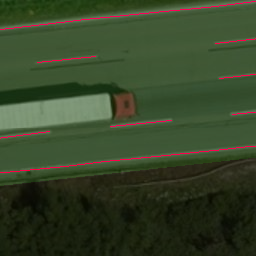}}}
		\resizebox{0.12\textwidth}{!}{\rotatebox{0}{
				\includegraphics{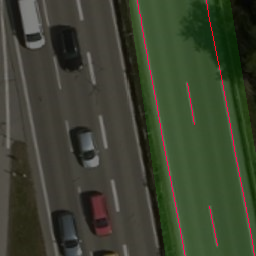}}}
		\resizebox{0.12\textwidth}{!}{\rotatebox{0}{
				\includegraphics{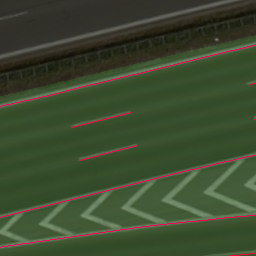}}}
		\\
	\end{tabular}
	}
	\caption{Satellite tile images fused with ground truth lane boundary geometry at tile pixel $[557869, 363521]$, $[557909, 363513]$, $[558036, 363507]$ and $[557879, 363518]$ at level $20$ from left to right. Road region highlighted in green, bounded by road boundaries. Lane marking highlighted in red.}
	\label{fig:road_model}
\end{figure}

The objective of this step is to build a classifier that can determine whether an image patch contains any lane marking pixel. Even though the ground truth road model is organized in chunk-wise structure~\footnote{Divide road along centerline into pieces evenly}, due to the specificity of the tile system, generating training patches in chunks unnecessarily queries the image server twice (each tile always contains two to three chunks at high resolution level). Hence, our solution is designed as tile-wise - for each tile along trajectory, project all control points of each functional line and connect them. To reduce noise (i.e. lane marking pixel of adjacent road surfaces), the surface region is bounded by road boundaries (given in the ground truth dataset). Samples contain lane marking pixels in red and road surface region in green are illustrated in Figure~\ref{fig:road_model}.

\begin{figure}[H]
		\begin{tabular}{cc}
			\resizebox{0.22\textwidth}{!}{\rotatebox{0}{
					\includegraphics{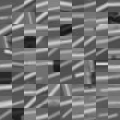}}}
			&
			\resizebox{0.22\textwidth}{!}{\rotatebox{0}{
					\includegraphics{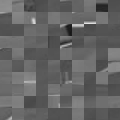}}}
			\\
			(a) & (b)
		\end{tabular}
		\caption{100 sample patches of original orientation lane-marking patches (a) and non-lane-marking patches (b).}
		\label{fig:patchmatrix}
\end{figure}

A sliding window is designed to crop training patches from corresponding satellite image within the road surface. The label for each patch is determined by whether there is any lane marking pixel hit in the current patch. To reduce misleading ground truth patches (the patch contains two independent lines), an appropriate window size should be thinner than lane width in real scale, which is $3.5$ meters~\footnote{New Guidelines for the Geometric Design of Rural Roads in Germany}. In this project, the ground resolution is approximately $0.15$ meter per pixel at tile level $20$, which means the patch size should be less than $24$. Examples of positive (contain lane marking pixel) and negative patches are shown in Figure~\ref{fig:patchmatrix}.

\begin{figure}[h]
	\centerline{
	\begin{tabular}{cc}
		\resizebox{0.24\textwidth}{!}{\rotatebox{0}{
				\includegraphics{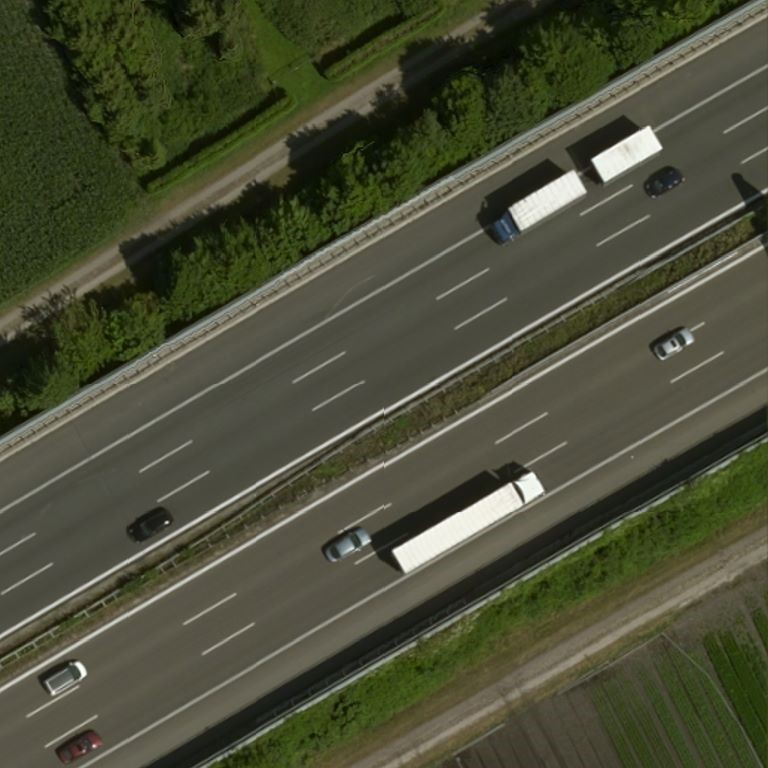}}}
		&
		\resizebox{0.24\textwidth}{!}{\rotatebox{0}{
				\includegraphics{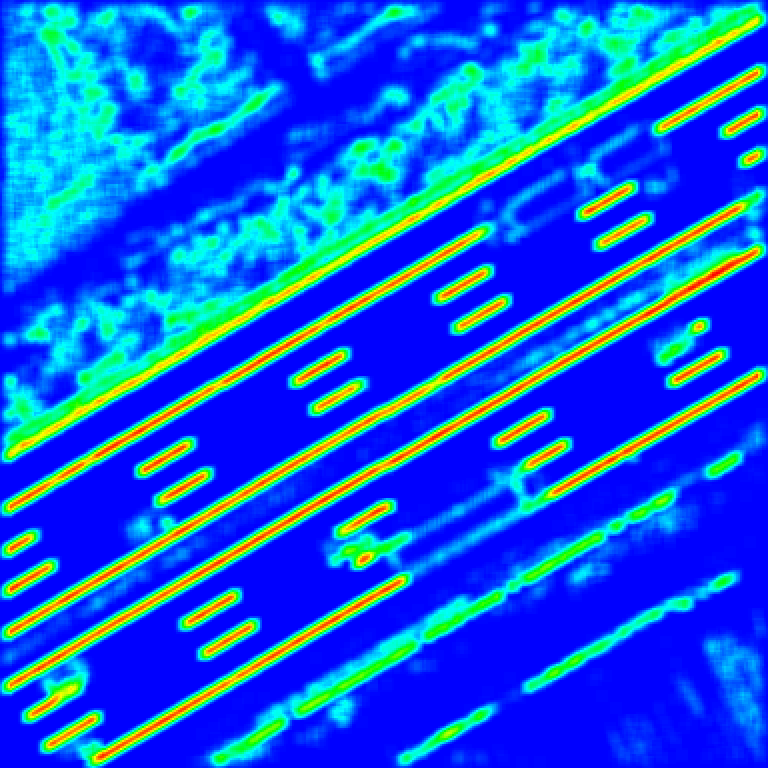}}}
		\\
		(a) & (b)
		\\
	\end{tabular}
	}
	\caption{Original satellite tile image (a) and its patch level lane marking probability map (b) at location  $48.2203\degree,11.5126\degree$.}
	\label{fig:probability_map}
\end{figure}

Given a satellite tile image (shown in Figure~\ref{fig:probability_map}~(a)), its corresponding probability map of patch level lane marking with certain configuration (patch size is $12$ pixels, use pixel representation feature and Random Forest classifier) is illustrated in Figure~\ref{fig:probability_map}~(b).

\subsection{Pixel-wise Lane Marking Segmentation}
Our patch level classification returns a probability map as output, which contains high probability lane marking regions like the red area shown in Figure~\ref{fig:probability_map}(b). However, since each region is always several pixels wide (depending on patch size and step length), this wide range obviously could not meet the requirement considering the definition of HD Maps~\cite{bittel2015pixel}. 

\begin{figure}[h]
		\begin{tabular}{c}
			\resizebox{0.46\textwidth}{!}{\rotatebox{90}{
					\includegraphics{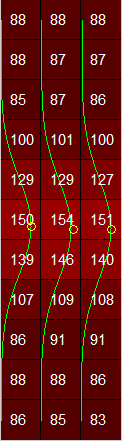}}}
			\\
		\end{tabular}
		\caption{Pixel-wise and sub pixel-wise lane marking segmentation visualization, each number inside pixel represents its intensity value converted from the raw RGB image.}
		\label{fig:sub_pixel_wise_segmentation}
\end{figure}

To segment and locate precise lane marking pixels, we consider pixels with the highest intensity in each slice of lane marking region perpendicular to road trajectory as lane-marking candidate. Then, we fit a line segment through the lane marking pixel candidates. For example, in Figure~\ref{fig:sub_pixel_wise_segmentation}, assuming the trajectory is \textbf{up}, for each row of this region, the highest intensity points are $151$, $154$, and $150$, respectively, which means the lane marking line segment should be the centerline of this region.

\begin{figure}[H]
		\begin{tabular}{c}
			\resizebox{0.48\textwidth}{!}{\rotatebox{0}{
					\includegraphics{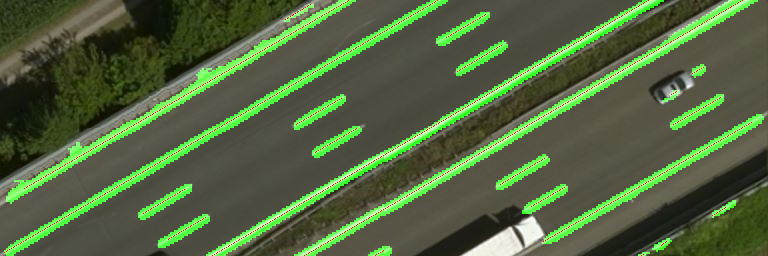}}}
			\\
			\resizebox{0.48\textwidth}{!}{\rotatebox{0}{
					\includegraphics{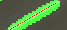}}}
			\\
		\end{tabular}
		\caption{Lane marking region candidates (green regions) and pixel-wise lane marking pixel candidates (red dots) overview (top) and zoom view (bottom).}
		\label{fig:line_candidate}
\end{figure}

Even though the satellite image resolution has already matched the lane marking width, limited by image compression, hardware constraints(lens, CMOS), and optical limitations (i.e. angular resolution), tiny/thin object will always be blurred at its boundaries when captured. Furthermore, to segment more precise lane marking pixel locations, we introduce sub pixel-wise segmentation. For each slice of the lane marking region, fit a Gaussian model (green lines in Figure~\ref{fig:sub_pixel_wise_segmentation}) and find the peak of each model (yellow circles in Figure~\ref{fig:sub_pixel_wise_segmentation}). Then, the lane marking pixel location becomes sub-pixel-wise instead of the naive pixel-wise of each slice of the lane marking region. Theoretically, line accuracy can be improved by at most half a pixel. The pixel-wise lane marking segmentation result is illustrated in Figure~\ref{fig:line_candidate}.

\subsection{Line Candidates Grouping, Classification and Linking}\label{sec:linking}

\begin{figure}[h]
		\begin{tabular}{cc}
			\resizebox{0.22\textwidth}{!}{\rotatebox{0}{
					\includegraphics{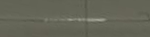}}}
			&
			\resizebox{0.22\textwidth}{!}{\rotatebox{0}{
					\includegraphics{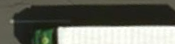}}}
			\\
			(a)
			\\
			\resizebox{0.22\textwidth}{!}{\rotatebox{0}{
					\includegraphics{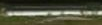}}}
			&
			\resizebox{0.22\textwidth}{!}{\rotatebox{0}{
					\includegraphics{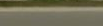}}}			
			\\
			(b)
			\\
		\end{tabular}
		\caption{Phenomenons cause mis-detection: bad painting quality and shadow (a), misleading objects that cause false positive: high reflective metal guardrail and cement curb (b).}
		\label{fig:good_bad}
\end{figure}

The previous step generates unstructured line segments without relative position and function label (solid/dashed line). Because of occlusion (i.e. trees, vehicles, buildings, and their shadows) and poorly painted lane markings(examples shown in Figure~\ref{fig:good_bad}~(a)), less true lane marking lines will be detected, while more misleading lines (false positive) will be detected if lane-marking-like objects appear (i.e. guardrail, curb, wall shown in Figure~\ref{fig:good_bad}~(b)). 

The method of of transforming the unstructured lines to structured lines with function labels contains three steps: grouping line candidates from each chunk, classify the function of each line group, and link the missing lines.

\begin{figure}[h]
		\begin{tabular}{cc}
			\resizebox{0.23\textwidth}{!}{\rotatebox{0}{
					\includegraphics{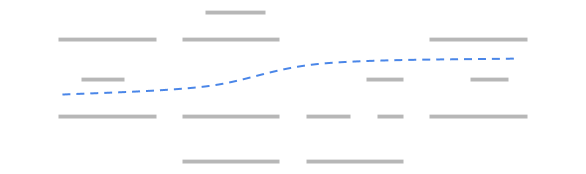}}}
			&
			\resizebox{0.23\textwidth}{!}{\rotatebox{0}{
					\includegraphics{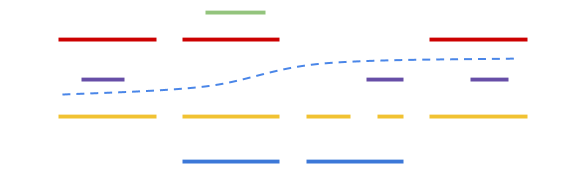}}}		
			\\
			(a) & (b)
			\\
			\resizebox{0.23\textwidth}{!}{\rotatebox{0}{
					\includegraphics{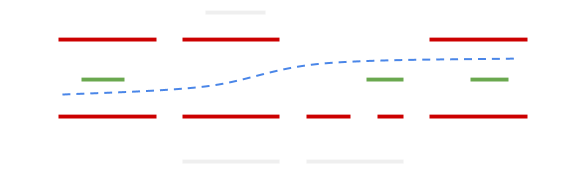}}}	
			&
			\resizebox{0.23\textwidth}{!}{\rotatebox{0}{
					\includegraphics{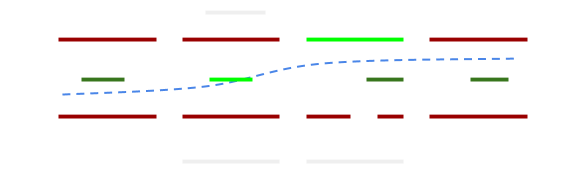}}}	
			\\
			(c) & (d)
			\\
		\end{tabular}
		\caption{Illustrations of Line Candidates grouping, classification and linking steps.}
		\label{fig:modeling}
\end{figure}

In the grouping step, whether or not to push a line into a group depends on the relative distance \footnote{There is no definition of line segment to line segment distance in a 2-D plane if they are not parallel, the relative distance here is the average distance between each point from one line segment to the other.} between the current line to all other line candidates in the current chunk, in the neighboring chunk(s), and their relative distances to road centerline/vehicle trajectory. For example, line candidates (gray) from four continuous chunks and vehicle trajectory (blue dashed) are illustrated in Figure~\ref{fig:modeling}~(a). After the grouping step, five groups are generated and colored in Figure~\ref{fig:modeling}~(b). On a certain portion of the road, for each group, the function label is determined by the ratio of the total length of detected line segments belong to this group, to the total length of road contains this line group. Typically, consider reasonable mis-detection and wrong detection, the length ratio of dashed line is below $40\%$ and the ratio of solid line is above $80\%$. In the task of modeling highway roads, there is an assumption that each road portion can have at most two solid lines bounding the (drivable) road surface. Figure~\ref{fig:modeling}~(c) illustrates the groups after the classification step, solid lines and dashed lines are colored in dark red and dark green, respectively, lines out of solid lines (drivable road surface) are colored in gray and will be ignored. In the final step, if one chunk does not contain a line that belongs to the group which passes this chunk, a synthetic line will be interpolated (light green lines shown in Figure~\ref{fig:modeling}~(d)).

\begin{figure}[h]
		\begin{tabular}{c}
			\resizebox{0.45\textwidth}{!}{\rotatebox{0}{
					\includegraphics{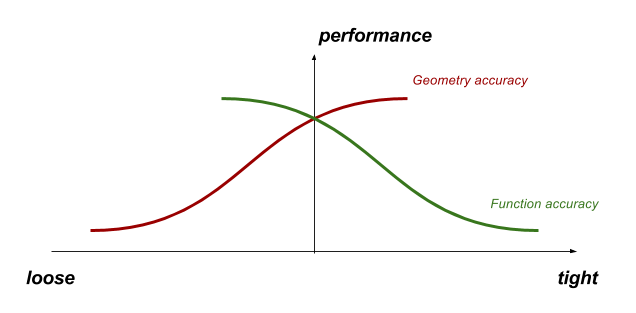}}}
			\\
		\end{tabular}
		\caption{Constrains tightness to modeling performance chart.}
		\label{fig:tightness_to_performance}
\end{figure}

Notice that in this grouping, classification, and linking procedure, numerous thresholds and constrains (i.e. distance threshold, search range, etc.) are needed to control the process. Generally speaking, if we abstract all these variables into one degree: loose (longer search range, wider distance threshold) and tight (shorter search range, narrower distance threshold) to reflect the abstract performance of the model, the tightness-to-performance chart is illustrated in Figure~\ref{fig:tightness_to_performance}\footnote{This chart is subject to mis-detection rate and wrong detection rate for the portion of road}. As we can see, it is a trade off between function accuracy and geometry accuracy \footnote{The definitions of function and geometry are described in Section~\ref{hd_maps_accuracy}}.

Furthermore, if any constraint (additional information) is provided when extracting the lane boundary geometry - for example, if we know the number of lanes for a certain portion of the road - this process can be fine tuned to generate a much better result.

\section{Lane Boundary Ground Truth Collection}
In paper~\cite{seo2012ortho}, the author uses manually-drew, pixel-based ground truth, represented in 'mask' format~\cite{seo2012augmenting}, to evaluate his accuracy. The author does not have detailed description and statistic of his dataset. The number of ground truth masks is 50, which also limits its persuasion.

To code our lane boundary geometry dataset, we built an interactive tool which allows us to manually draw lane boundaries from scratch and present background image from various sources (i.e. point cloud projection, satellite imagery, etc.). The user interface is shown in Figure~\ref{fig:code_rule}. An lane boundary geometry extraction from LiDAR point cloud pipeline will be executed at the very beginning to generate near-perfect lane boundaries to boost our modeling efficiency from $29.2$ meters per minute to $12.8$ meters per minute\footnote{Time efficiency is dependent on the number of lanes and the road structure. In our dataset, the majority of lane number is 3.}. Then we use our tool to edit (delete, move, insert, etc.) the control points on lane boundary lines to make align them with the background imagery perfectly. 

As one of the contributions, we will publish this line based lane boundary geometry ground truth on our FTP server once the paper is accepted. In this section, we are going to discuss the road selection, coding rules, lane boundary geometry representation, and potential system errors. More lane boundary geometry data of diversified scenarios (luminance condition, country, etc.) will be published as future work.

\subsection{Lane Boundary Data Description}\label{sec:code_rule}
We collect lane boundary geometry on Germany Autobahn A99 (Bundesautobahn 99), from location $48.2206\degree 11.5153\degree$ to $48.2057\degree 11.4586\degree$, divided into seven portions (five for training and two for testing) to exclude overpass structures and other unexpected scenarios.

\begin{figure*}[h] 
		\begin{tabular}{c}
			\resizebox{0.9\textwidth}{!}{\rotatebox{0}{
					\includegraphics{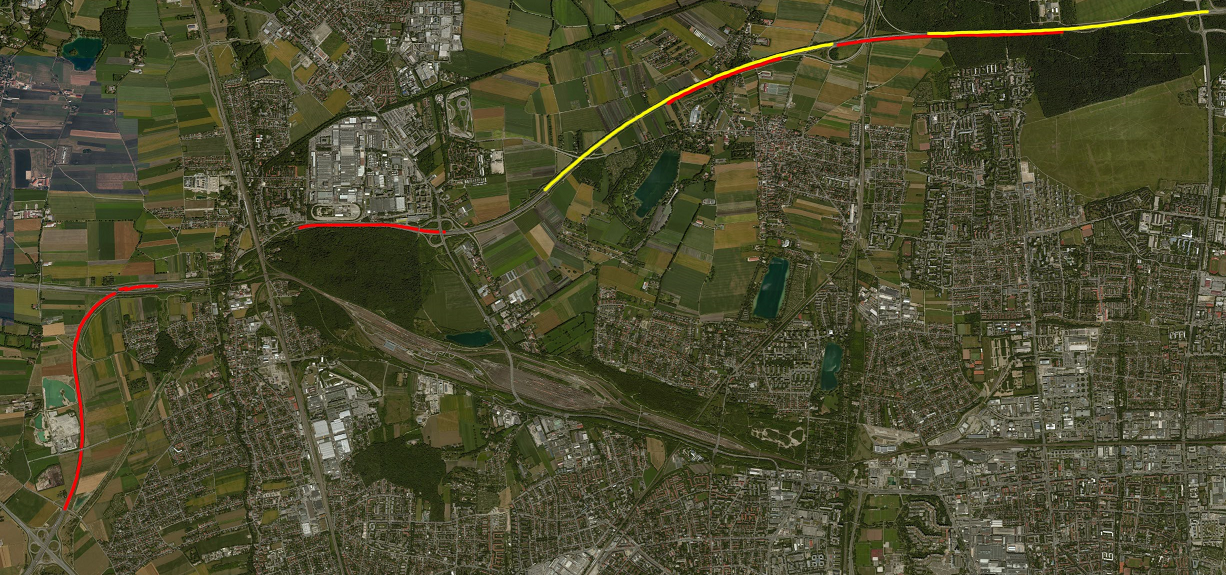}}}
			\\
		\end{tabular}
		\caption{Illustration of our ground truth portions highlighted on aerial imagery, red for training and yellow for testing.}
		\label{fig:germany_a99}
\end{figure*}

\begin{figure}[H]
		\begin{tabular}{cc}
			\resizebox{0.22\textwidth}{!}{\rotatebox{0}{
					\includegraphics{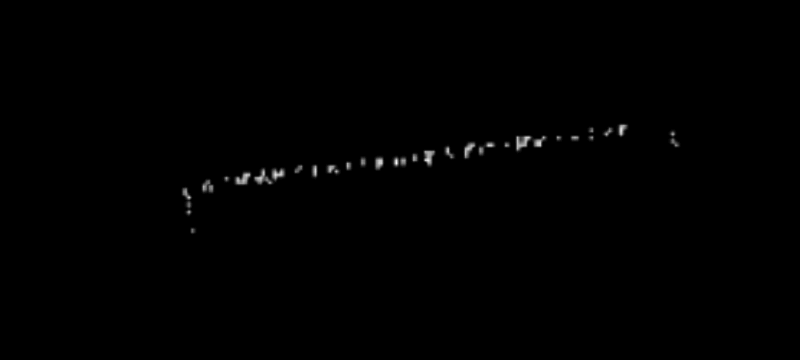}}}
			&
			\resizebox{0.22\textwidth}{!}{\rotatebox{0}{
					\includegraphics{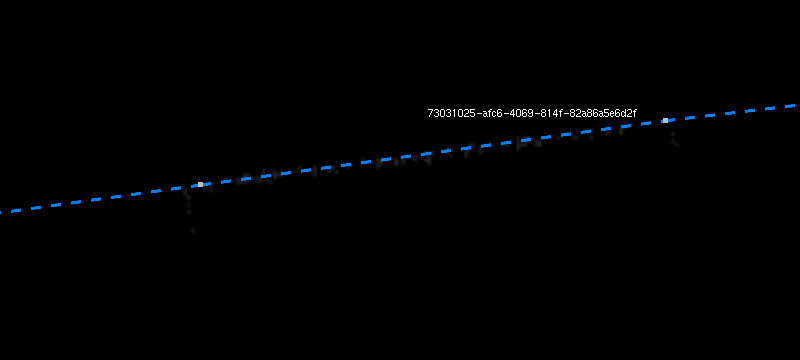}}}
			\\
			(a) & (b)
			\\
			\resizebox{0.22\textwidth}{!}{\rotatebox{0}{
					\includegraphics{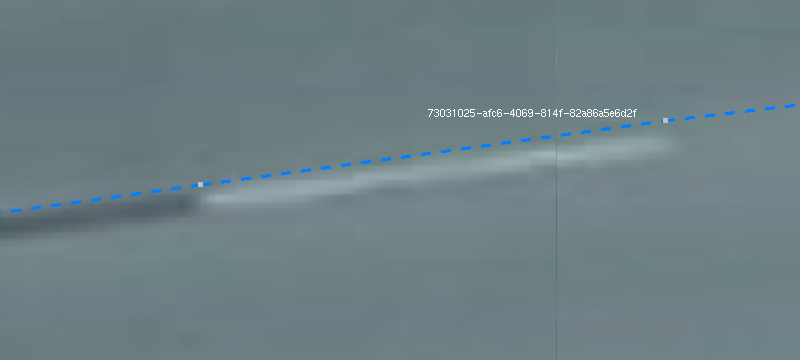}}}
			&
			\resizebox{0.22\textwidth}{!}{\rotatebox{0}{
					\includegraphics{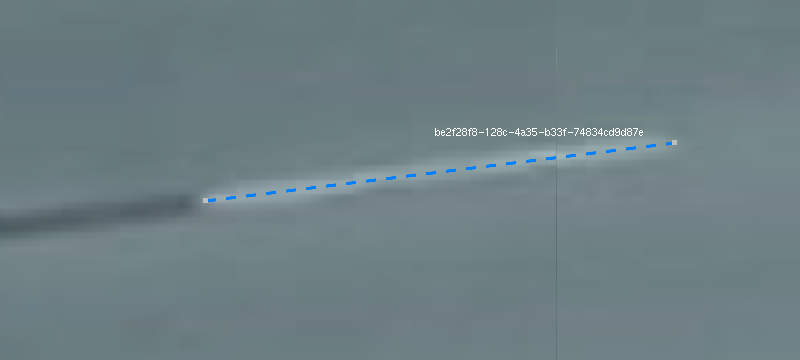}}}
			\\
			\\
			(c) & (d)
			\\
		\end{tabular}
		\caption{Lane boundary line in point cloud view (a), test lane boundary line (continuous dashed line) in point cloud view (b), test lane boundary line in satellite imagery mode (misalignment between two systems can be obtained)(c) and train lane boundary line (isolated dashed line) in satellite imagery view (d). UUID denotes the line id.}
		\label{fig:code_rule}
\end{figure}

This dataset contains training data (approximately $10.14$ kilometers) and testing data ($6.07$ kilometers), which follows similar coding rules but two key differences in coding the dashed line and the coordinate system due to the particularity of this research. In training set (coverage shown in Figure~\ref{fig:germany_a99}, highlighted in red), dashed lines are represented as isolated line segments (two end points), shown in Figure~\ref{fig:code_rule}(b), and aligned with satellite imagery coordinates. In testing dataset (coverage shown in Figure~\ref{fig:germany_a99}, highlighted in yellow), dashed lane markings belong to one set of continuous/sequential control points if they are semantically treated as one line and aligned with point cloud, as shown in Figure~\ref{fig:code_rule}(c). All control points are placed right in the middle of their corresponding lane marking.

Also, in this dataset, we code road boundaries (i.e. guardrail, curb) and use them to separate road surface from non-road surface (to exclude lane-marking like objects outside the road surface).

\subsection{Data Annotation}
To represent lane boundary geometry and lane boundary lines for user convenience, lane boundary lines are evenly divided into $12$-meter chunks. Each chunk is wrapped up in a single JSON file that follows this structure\footnote{Map version specifies the map version which ground truth lane boundary lines aligns to, follows Bing Maps URL http://a0.ortho.tiles.virtualearth.net/tiles/a/[quadkey].jpeg?g=[map version]}:

\fbox{
	\begin{minipage}{25em}
		Chunk JSON file\\
		\hspace*{12pt}id: INT		\\
		\hspace*{12pt}map version: INT \\
		\hspace*{12pt}lines:			\\
		\hspace*{12pt}\hspace*{12pt}line id:  STRING: UUID\\
		\hspace*{12pt}\hspace*{12pt}type:     STRING: ['solid'/'dashed'/'trajectory']\\
		\hspace*{12pt}\hspace*{12pt}points:   FLOAT: [n by (latitude, longitude)] matrix
	\end{minipage}}

\subsection{Errors in latitude, longitude, and altitude}
Aerial imagery and point cloud are stored/represented in two coordinate systems - Mercator projection coordinate system~\cite{bingmap} and Cartesian coordinate system~\cite{atanacio2011lidar,schwarz2010lidar} - because of their acquisition techniques. Aerial image tile system is designed by demand years ago, but it inevitably has heavy distortion.

To avoid distortion, point cloud processing procedures and our labeling tool are designed to process data in Cartesian coordinate system. The different coordinate systems lead to a slight distance error when coded on these two layers - point cloud representation in a Cartesian local tangent plane: North-East-Up (NEU) and imagery represented in Mercator projection. Assume location $[\phi, \lambda, 0]$ at zoom level $l$, the Euclidean distance $d_p(\phi, \lambda, d_x, d_y)$ between the points back-projected through Mercator projection and Cartesian coordinate transformations of pixel shift $d_x, d_y$ is complicated.

%
%
%
%
%
%

\begin{figure}[H]
	\begin{tabular}{c}
		\resizebox{0.45\textwidth}{!}{\rotatebox{0}{
				\includegraphics{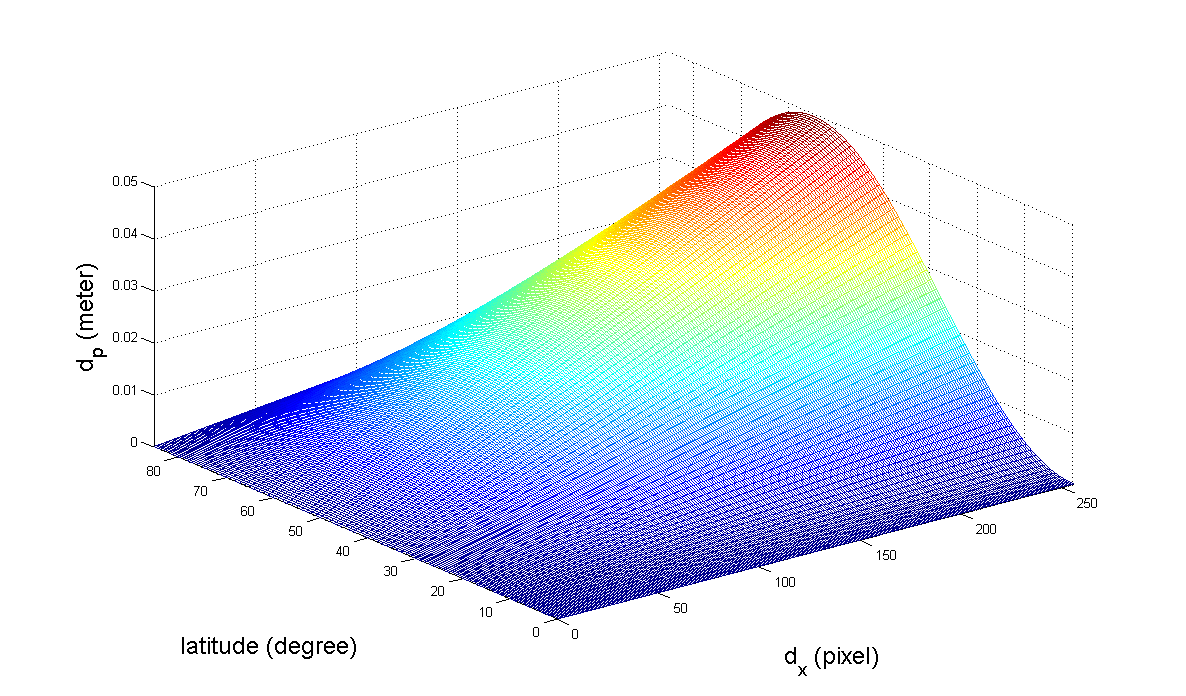}}}
		\\
	\end{tabular}
	\caption{Illustration of $d_p(\phi, d_x, d_y, l)$ when $l = 20$ and $d_y = 0$.}
	\label{fig:error_between_2_coordinates}
\end{figure}

After simplification, $d_p$ can be represented as the function of latitude $\phi$, pixel shift $[d_x, dy]$, and zoom level $l$, and the illustration of function $d_p(\phi, d_x, d_y, l)$ at certain zoom level $l = 20$ and certain $dy = 0$ is shown in Figure~\ref{fig:error_between_2_coordinates}. According to precision requirements from most 'HD' definitions~\cite{herehd, tomtomhd, baiduhd}, the error caused by fusion of two coordinate systems (less than $5$ cm all around the world at tile level $20$) does not have an impact on the final accuracy.

\section{Experiments}\label{sec:experiments}
In the proposed methodology, we test numerous patch configurations (patch feature, size) and machine learning techniques to find the appropriate classifier and run our end-to-end program to extract lane boundary geometry using this optimum classifier. In this section, we are going to present both patch level accuracy and final extracted model accuracy by using the metrics described in the next sub-section.

\subsection{Accuracy Definition}\label{hd_maps_accuracy}
Lane boundaries is a collection of numeric and functional polylines/splines in control point representation~\cite{schindler2012generation}, we cannot simply report the accuracy in pixel-wise representation as~\cite{seo2012ortho} proposes.

Considering the features of lane boundary and the misalignment between ground truth and modeling coordinates, we propose two metrics for a persuasive performance score: \textbf{function level} and \textbf{geometry level} accuracy. Theoretically, the misalignment between two coordinates causes transformation from one model to another to shift, scale, rotate, and even skew. In our task, scaling, rotating, and skewness are unnoticeable so they can be ignored, only shift transformation will be considered.

Given ground truth lines $L_{i} = \lbrace l_{i,1}, l_{i,2},...., l_{i,n} \rbrace$ and predicted lines $L'_{i} = \lbrace l'_{i,1}, l'_{i,2},...., l'_{i,m} \rbrace$ from $i$th chunk, the first step is to match $l \in L_{i}$ and $l' \in L'_{i}$. Let $d(l, l')$ denotes the distance between pair lines $l$ and $l'$ with sign (for example, left for negative) and $Pair(l_i, L'_i)$ denotes the paired line in $L'_i$ of $l_i$. Align two models by using each $l_i$ and each $l'_i$, find the 

\begin{eqnarray*}
	\operatorname{argmin}_{all\:pairs}(\sum_{l_{i,j} \in L_i}(d(l_{i,j}, Pair(l_{i,j}, L'_i))))
\end{eqnarray*}

\begin{figure}[h]
	\begin{tabular}{cc}
		\resizebox{0.45\textwidth}{!}{\rotatebox{0}{
				\includegraphics{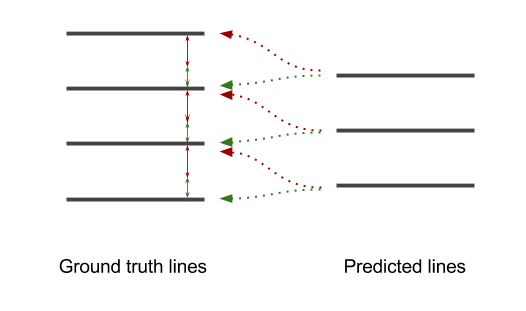}}}
		\\
	\end{tabular}
	\caption{Ground truth lane boundary lines (left) and predicted lane boundary lines (road) pairing.}
	\label{fig:gt_result_pairing}
\end{figure}

 Figure~\ref{fig:gt_result_pairing} shows two pairs: green dashed arrow and red dashed arrow. By comparing the total length of the green and red solid arrows, the best match is green dashed arrow. Given distance threshold $T_d$ from requirement (for example, HD Maps), a correct detection of $l'_i$ and $l_i$ is counted if their functions are matched and $d(l'_i, l_i) < T_d$. Then, the accuracy of the predicted model compared to ground truth can be represented in 

\begin{eqnarray*}
	&\operatorname{\textit{precision}}_{\operatorname{function}} = \frac{number\:of\:correct\:detections}{(\parallel\textbf{L} \parallel)}\\
	&\operatorname{\textit{recall}}_{\operatorname{function}} = \frac{number\:of\:correct\:detections}{(\parallel \textbf{L}' \parallel)}\\
	&\textbf{L} = \lbrace L_i \rbrace, \textbf{L}' = \lbrace L'_i \rbrace, i \in road
\end{eqnarray*}

To calculate geometry accuracy, for each pair $\lbrace l_{i,j}:Pair(l_{i,j},L'_i) \rbrace$ in $L_i$, the $\operatorname{\textit{shift}}$ is defined as $AVG_{i}(d(l_i:Pair(l_{i,j},L'_i))), L'_i\:and\:L_i \in \textbf{L}$, and the $\operatorname{\textit{performance}}_{geometry}$ is defined as $AVG_i(1 - \frac{\sigma(d(l_i:Pair(l_{i,j},L'_i)))}{\sigma_{max}(\parallel L_i \parallel)})$, while $\sigma_{max} = \sigma(L_{max})$ is used to normalized precision for each chunk, where $L_{max} = M \cup N$, set $M$ contains $\operatorname{\textit{floor}}(\frac{\parallel L_i \parallel)}{2})$'s $T_d$ and set $N$ contains $\operatorname{\textit{ceiling}}(\frac{\parallel L_i \parallel)}{2})$'s  $-T_d$ (for example, if $\parallel L_i \parallel)$ equals $4$, then $\sigma_{max}(\parallel L_i \parallel) = \sigma([T_d, T_d, -T_d, -T_d])$). With this $\{\operatorname{\textit{perfomance}}:\operatorname{\textit{shift}}\}$ metric, we can present the lane boundary geometry reasonably if the alignment between two source coordinates is unknown. Also, we can tweak the parameters of end-to-end program to generate expected model depends on the project requirement. For example, if alignment is not a requirement and we want to address the LiDAR shadow issue from point cloud, we would need to tweak the configurations with the lowest \textit{performance}.

\subsection{Patch level accuracy}\label{sec:patch_level_accuracy}
\begin{figure}[h]
	\begin{tabular}{c}
		\resizebox{0.48\textwidth}{!}{\rotatebox{0}{
				\includegraphics{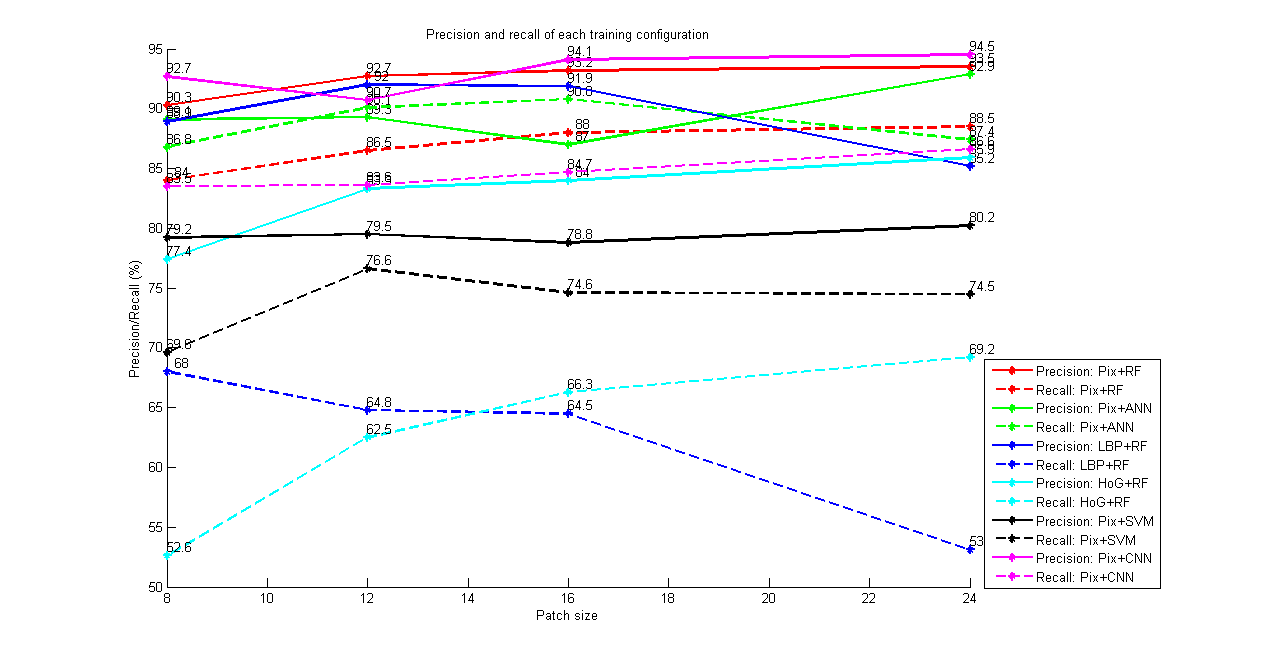}}}
		\\
	\end{tabular}
	\caption{Patch level precision and recall of each training configuration.}
	\label{fig:patch_level_precision_recall}
\end{figure}

To find the best patch level classifier, we crop tile image with given ground truth data as described in Section~\ref{tile_and_patch}, with configurable variables such as as patch size ($8, 12, 16, 24$), patch feature (pixel representation and gradient based features such as Histogram of Oriented Gradients (HOG), Local Binary Pattern (LBP)), and numerous machine learning techniques (Random Forest (RF), Support Vector Machine (SVM), Artificial Neural Network (ANN) and Convolutional Neural Network (CNN)) to evaluate their precision and recall. Figure~\ref{fig:patch_level_precision_recall} shows the performances of different training configurations of $10$-fold Cross Validation.

According to our evaluation result, the change in patch size does not have a big impact on patch level performance. Considering the computational cost, a patch size of $12$ is used in our final patch level classifier to generate a dense/smooth lane marking probability map shown in Figure~\ref{fig:probability_map}~(b).

\subsection{Lane Boundary Geometry Accuracy}\label{sec:model_performance}
With the pre-trained classifier and our end-to-end solution, we tweaked parameters and thresholds as mentioned in Section~\ref{sec:linking} to evaluate the performance of our approach on the testing set. The function level and geometry performances are shown in Table~\ref{tbl:model_level_precision_recall} below.

\begin{table}[h]
  \caption{Model level performance with different conditions.}
  \label{tbl:model_level_precision_recall}
  \includegraphics[width=\linewidth]{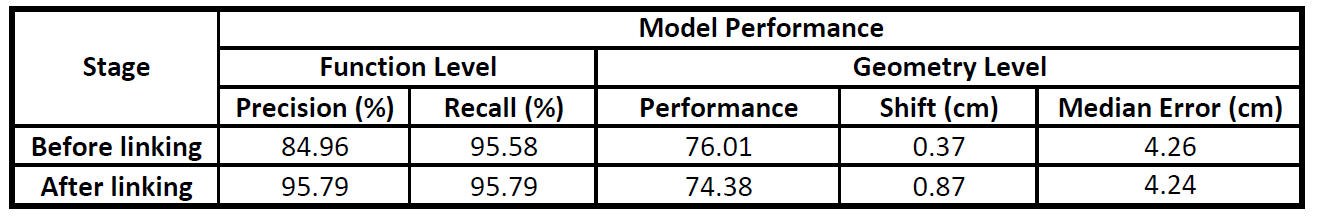}
\end{table}

\begin{figure}[h]
	\begin{tabular}{c}
		\resizebox{0.45\textwidth}{!}{\rotatebox{0}{
				\includegraphics{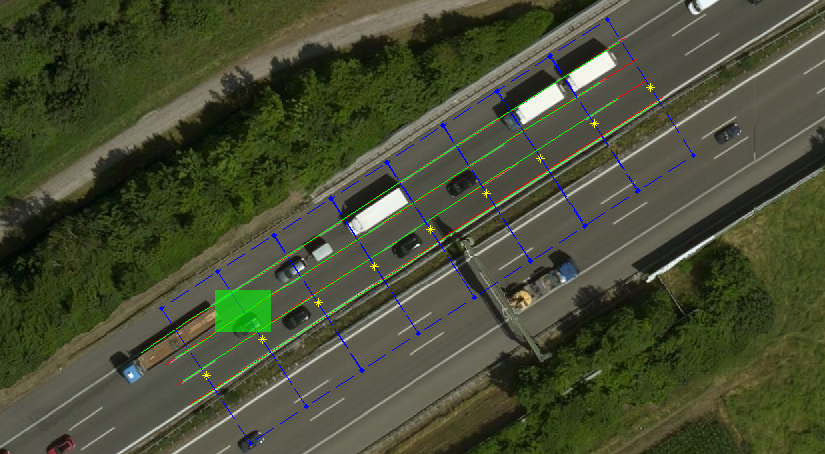}}}
		\\
		\resizebox{0.45\textwidth}{!}{\rotatebox{0}{
				\includegraphics{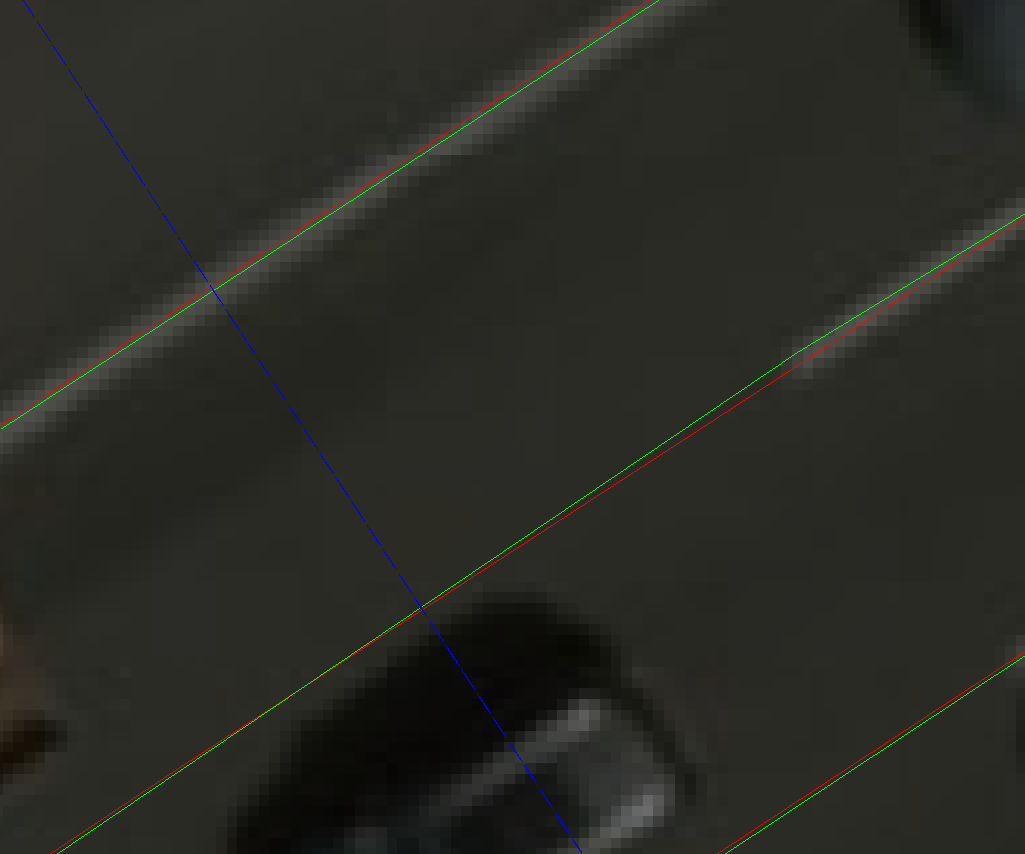}}}
	\end{tabular}
	\caption{Result visualization overview (top) and zoom view of the green box (bottom).}
	\label{fig:gt_result}
\end{figure}

Sample of ground truth lane boundaries and final extracted lane boundaries rendered on satellite imagery is shown in Figure~\ref{fig:gt_result}~(a). Each chunk is bounded by blue rectangle, yellow stars denote road trajectory, ground truth model is rendered in red, and the resulting lane boundary geometry is rendered in green.

The \textit{shift} from experiment result shows the misalignment between ground truth coordinate (from point cloud) and testing data coordinate (from satellite imagery) is negligible on testing data, which was also validated by our observation. Benefited from the above-average conditions of the testing road (good painting quality, light traffic), the extracted lane boundaries achieved impressive results when measured against the ground truth before and after the linking stage at geometry level. Function level precision improves $10.83\%$ by the linking stage while geometry level performance only dropped $1.63\%$ due to the interpolated, synthetic lane boundaries. Even though the median error of the results is lower than $5$ cm, limited by original satellite imagery resolution, we can claim that our lane boundary geometry accuracy is $30$ cm~\footnote{https://blogs.bing.com/maps/2011/06/27/bing\--maps\--unveils\--exclusive\--high\--res\--imagery\--with\--global\--ortho\--project/}.

\section{Concluding Remarks and Future Directions}

In this paper we present a novel approach to automatically extract lane boundary geometry from reasonably high resolution satellite imagery. It complements the existing ground-level based methods with advantages such as cost-effectiveness, wider coverage, and better updatability. We also designed a comprehensive lane boundary geometry evaluation metric and published our lane boundary geometry dataset.


The following areas will be investigated for further improvement:

\begin{enumerate}

\item Elevation information is not available in satellite imagery. This can be solved with additional data source such as DEM and the High Definition Digital Elevation Model (HD DEM) database~\cite{andi2017ssdbm}.


\item We will combine both ground level and airborne data to extract lane boundary geometry. Alignment of these two data sets is critical for the fusion. Some feature points visible in both model could register the two data sets. For example, road surface markings and pole like objects are good control point candidates.


\item By adding absolute/relative spatial information to each pixel~\cite{pinheiro2014recurrent}, a machine learning based (especially Recurrent Convolutional Neural Networks) lane marking pixel classification may substitute the approach of combining machine learning based lane marking patch classification and pixel-wise segmentation proposed in this paper.


\item Other than limited access roads such as highway and expressway, there is a large number of urban road networks that also need to be modeled in high definition for fully automated driving. Occlusions and shadows of urban roads in satellite imagery make it very challenging to apply our approach. In addition, the complex geometry and topology of urban road may require a different lane model for automation.

\end{enumerate}

\bibliographystyle{ACM-Reference-Format}
\bibliography{road_modeling}

\end{document}